\documentclass[runningheads]{llncs}
\usepackage[T1]{fontenc}
\usepackage{graphicx,verbatim}
\usepackage{tablefootnote}

\usepackage{caption}
\usepackage{subcaption}
\usepackage{booktabs}
\usepackage{multirow}
\usepackage{graphicx}
\usepackage{hyperref}
\usepackage[symbol]{footmisc}

\begin{document}
\title{Improving Pre-trained Adult Glioma Segmentation Models Using only Post-processing Techniques}

%

\author{ Abhijeet Parida\inst{1,2}*, Daniel Capell\'{a}n-Mart\'{i}n\inst{1,2}*, Zhifan Jiang\inst{1}* \\ Nishad Kulkarni\inst{1} , Krithika Iyer\inst{1}, Austin Tapp\inst{1}, Syed Muhammad Anwar\inst{1,3}, \\Mar\'{i}a J. Ledesma-Carbayo\inst{2}  and Marius George Linguraru\inst{1,3}
\email{mlingura@childrensnational.org}}

\authorrunning{A. Parida, D. Capell\'{a}n-Mart\'{i}n, Z. Jiang  et al.}
%
\titlerunning{Improving Pre-trained Segmentation Models Using Post-Processing}
\institute{
Sheikh Zayed Institute for Pediatric Surgical Innovation, \\Children’s National Hospital, Washington, DC, USA 
\and 
Universidad Polit\'{e}cnica de Madrid and CIBER-BBN, ISCIII, Madrid, Spain
\and 
School of Medicine and Health Sciences, \\George Washington University, Washington, DC, USA
}

\maketitle              
\begin{abstract}
Gliomas are the most common malignant brain tumors in adults and are among the most lethal. Despite aggressive treatment, the median survival rate is less than 15 months. Accurate multiparametric MRI (mpMRI) tumor segmentation is critical for surgical planning, radiotherapy, and disease monitoring. While deep learning models have improved the accuracy of automated segmentation, large-scale pre-trained models generalize poorly and often underperform, producing systematic errors such as false positives, label swaps, and slice discontinuities in slices. These limitations are further compounded by unequal access to GPU resources and the growing environmental cost of large-scale model training. In this work, we propose adaptive post-processing techniques to refine the quality of glioma segmentations produced by large-scale pretrained models developed for various types of tumors. We demonstrated the techniques in multiple BraTS 2025 segmentation challenge tasks, with the ranking metric improving by 14.9 \% for the sub-Saharan Africa challenge and 0.9\% for the adult glioma challenge. This approach promotes a shift in brain tumor segmentation research from increasingly complex model architectures to efficient, clinically aligned post-processing strategies that are precise, computationally fair, and sustainable.

* These authors contributed equally.


\end{abstract}

\section{Introduction}

Gliomas are the most common malignant brain tumors in adults and remain among the deadliest among cancer types. Despite aggressive treatment strategies (maximal surgical resection, chemotherapy, radiotherapy), median overall survival duration is around 15 months. Malignant gliomas collectively account for approximately $2.5\%$ of all cancer-related deaths~\cite{sabouri2024survival}. 
Considering these alarming facts, there is a pressing need for tools that enable earlier detection, accurate characterization, and effective treatment planning to improve patient outcomes. Central to the clinical management of gliomas is multi-parametric magnetic resonance imaging (mpMRI), which provides non-invasive, high-resolution visualization of tumor anatomy and physiology. mpMRI–based tumor segmentation of tumors is routinely used to: (i) guide surgical resection margins~\cite{shaver2019optimizing}, (ii) paint radiotherapy dose~\cite{brighi2022repeatability}, and (iii) monitor volumetric disease progression~\cite{climent2025deep}. Despite its critical importance, tumor segmentation is still performed manually: an iterative and labor intensive process that can consume substantial expert time per case~\cite{ANANTHARAJAN2024101026}. In addition, the manual segmentation process is highly subjective, reducing the reproducibility between observers. Several deep learning models have recently been proposed to segment various pathologies~\cite{abidin2024recent,verma2025brain}. Automating the tumor segmentation step with deep learning models promises significant time savings and greater standardization across centers.

Since 2012, Brain Tumor Segmentation (BraTS) challenges, held in conjunction with the International Conference on Medical Image Computing and Computer Assisted Intervention (MICCAI), have become a leading contributor towards the development and benchmarking of automated tumor segmentation models by providing large, expertly annotated, and standardized datasets. The 2025 edition introduced the largest expert-annotated glioma data set and standardized lesion-wise normalized surface distance (NSD) metrics~\cite{6975210,brats2021}. During this period, the best performing methods have primarily relied on the nnU-Net framework~\cite{nnunet}, which extends the original U-Net architecture~\cite{unet} with modality- and dataset-specific adaptations to deliver state-of-the-art performance for 3D biomedical image tasks. In contrast, large-scale pre-trained models underperform in glioma segmentation and may perform inconsistently and generalize poorly even after light fine-tuning~\cite{foundation}. Their residual errors (tiny false positive islands), segmentation label swaps, or slicewise discontinuities are systematic and easily correctable. Previous BraTS winners have shown that post-processing steps such as ensemble voting~\cite{maani2023advanced}, size-sensitive connected component filtering~\cite{capellan2023model}, ET-to-NCR relabeling heuristics~\cite{capellan2023model}, and adaptive refinement~\cite{parida2024adult,jiang2024magnetic} can significantly improve segmentation accuracy.  

Beyond the choice of model and post-processing algorithms, socioeconomic conditions also strongly affect the challenge outcomes. Access to GPUs remains highly unequal, with recent surveys showing that GPU resources are highly concentrated in high-income settings, leaving researchers, especially in low and middle-income countries, without adequate resources~\cite{kudiabor2024ai}. 
Moreover, training complex deep learning models on GPUs consumes considerable energy; therefore, the environmental impact of using GPUs is becoming impossible to ignore. 
However, post hoc refinement using simple image-processing techniques typically requires only a few CPU hours. Thus, adaptive post-processing offers a compute-democratic path, allowing all steps to run on commodity CPUs. This can broaden meaningful participation in BraTS while simultaneously shrinking the carbon footprint of medical-AI research.

Taken together, the saturation of performance gains in large-scale pre-trained segmentation model architectures, the unequal access to computational resources, and the environmental costs of large-scale training underscore the need for more efficient alternatives. Therefore, this article proposes samplewise adaptive post-processing techniques on foundation models for the segmentation of adult gliomas (\textbf{GLI}) and adult gliomas in sub-Saharan African patients (\textbf{SSA}). We show performance gains over the base foundation model without additional GPU training. We advocate a shift in brain MRI segmentation research from ever deeper backbones toward smarter, greener, and clinically aligned post-processing — promoting accuracy, computational equity, and sustainability.
\section{Previous Work}
\subsection{Glioma Segmentation Models}\label{sec:SM}

The \textit{BraTS orchestrator}~\cite{kofler2025brats} provides dockerized access to the best performing solutions from previous BraTS challenges, enabling fair, side-by-side evaluation of winning models. Since these models generalize well and have been trained on large datasets using the best practices to get the best performance, they are considered our large-scale pre-trained segmentation models. 
For pretreatment adult glioma (\textbf{GLI-pre}), the BraTS~2023 winner combined SwinUNETR~\cite{hatamizadeh2021swin} and nnU-Net~\cite{nnunet} in an ensemble trained on real and GAN-augmented data~\cite{ferreira2024we}.  
A closely related ensemble from the same group secured the first place in the post-treatment GLI (\textbf{GLI-post}) task at BraTS~2024.  

For adult glioma in sub-Saharan African patients (\textbf{SSA}), the leading model of BraTS~2024 first pre-trained on the large GLI-pre cohort and then fine-tuned on the modest SSA dataset, using a MedNeXT~\cite{mednext}–nnU-Net ensemble~\cite{parida2024adult}.  
Although these architectures differ in backbone details, their training recipes converge on heavy GPU usage, large-scale augmentation, and multifold ensembling to squeeze out the last increments in Dice.

\subsection{Adaptive Post-processing}

To mitigate systematic residual errors that survive end-to-end training, several groups have proposed \emph{adaptive post-processing}, i.e., selecting sample-specific refinements rather than applying a fixed post-processing pipeline to every case.  
Jiang \textit{et~al.}~\cite{jiang2024enhancing} formalized the concept and demonstrated that radiomic features of the predicted mask can guide which operations (connected-component filtering, ET$\rightarrow$NCR relabeling, etc.) maximize Dice gain for a given sample.  
Follow-up work clustered cases in radiomic space to learn cluster-wise thresholds for component removal~\cite{jiang2024magnetic}, and Parida \textit{et~al.}~\cite{parida2024adult} also applied similar heuristic rules to win the BraTS~2024 SSA challenge. These refinements run entirely on CPUs and boost generalization.

\section{Challenge \& Data Description}
\begin{figure}[ht]
    \centering
    \includegraphics[width=\linewidth]{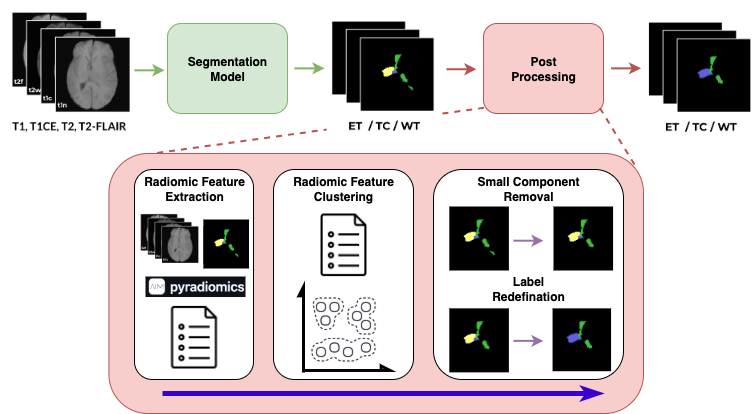}
    \caption{The process pipeline shows the high-level overview of our proposed post-processing pipeline. It shows the major steps of PyRadiomics-based feature extraction, clustering of samples based on the radiomic signature, removal of small components, and small label redefinition to get the final segmentation.}
    \label{fig:flowchart}
\end{figure}

\noindent\textbf{BraTS 2024 GLI}~\cite{de20242024,baid2021rsnaasnrmiccaibrats2021benchmark,bakas2017advancing} aims at automatic segmentation of diffuse gliomas on multi-institutional, clinically acquired pre and post-treatment mpMRI scans.  
Each case provides four sequences collected routinely: precontrast T1-weighted (T1), contrast-enhanced T1-weighted (T1CE), T2-weighted (T2), and T2-weighted fluid-attenuated inversion recovery (T2-FLAIR).  
The current version contains 1,251 GLI-pre and 1,350 GLI-post training cases. For the validation set, there are 219 GLI-pre and 188 GLI-post cases. Ground truth annotations cover six clinically relevant regions:
non-enhancing tumor core (NETC, label 1), surrounding non-enhancing FLAIR hyperintensity (SNFH, label 2), enhancing tumor tissue (ET, label 3), resection cavity (RC, label 4), the tumor core (TC = ET + NETC), and the whole tumor (WT = ET + SNFH + NETC).  
The challenge ranking is based on the Dice and the normalized surface distance (NSD) computed per lesion.

\noindent\textbf{BraTS 2024 SSA}~\cite{adewole2023brain,adewole2025brats} is the largest public collection of annotated pretreatment glioma scans from adult African patients, including low-grade gliomas and glioblastomas.  
It uses four MRI sequences (T1, T1CE, T2, T2-FLAIR) and the same label taxonomy (NETC, SNFH, ET, TC, WT) as the GLI task.  
The data set comprises 60 training and 35 validation cases; test labels are withheld.  
Evaluation employs identical Dice/NSD metrics.

\section{Methods}
\label{sec:methods}
We built on large-scale pre-trained segmentation networks and included a task specific, three stage post‑processing pipeline (Fig.~\ref{fig:flowchart}).  
The stages were  
(i)~\emph{radiomic feature extraction and case clustering},  
(ii)~\emph{thresholding to delete small, isolated components}, and (iii)~\emph{thresholding to fix label mix‑ups}.

We optimized the thresholds and chose the best model with the ranking approach proposed by the BraTS team LaBella \textit {et~al.}~\cite{LaBella2025-oh}. LaBella \textit {et~al.} stated - ``The evaluation is done in a hidden test set, computing lesion-wise metrics in all regions and comparing ranks of all the submissions rather than the metrics''. We folowed the idea and replicated this procedure, we built an internal ranking metric that produced a single score. The ranking metric was optimized for a lower score where a smaller value means better performance on Dice and NSD at the 0.5 and 1 mm thresholds for each of the regions. The code is available on GitHub \footnote[2]{\url{https://github.com/Pediatric-Accelerated-Intelligence-Lab/BraTS-Unofficial-Ranker}}.

The ranking metric approach was chosen because it was robust to outlier predictions. Further, the ranking metric allowed us to optimize for a single value while aligning with the contest evaluation pipeline. 

\noindent\textbf{Radiomic Feature Extraction and Clustering:}
For the prediction of the WT ensemble, we calculated 386 radiomic features using \textit{PyRadiomics}~\cite{van2017computational}, following the protocol of Jiang \textit{et~al.}~\cite{jiang2023automatic}.  
Therefore, each case had- 14 shape descriptors that capture tumor geometry, and  93 intensity \& texture descriptors- for each of the four MRI sequences (T1, T1Gd, T2, FLAIR).  
Using principal component analysis (PCA), we retained the principal components that explain $90\,\%$ of the variance and then partitioned the cases with $k$ means clustering~\cite{jiang2024enhancing,jiang2024magnetic}.  
The optimal number of clusters was determined by maximizing the silhouette coefficient in the training folds.  
Each new case was assigned to the nearest cluster at test time, allowing for the post-processing to be applied on the basis of its radiomic signature. 

\noindent\textbf{Threshold identification for removing small components ($p_{cc}$):} Within each cluster and for each label (NETC, SNFH, ET, and RC), we performed a grid search over minimum size thresholds to retain the lesion~\cite{parida2024adult}. Each threshold was evaluated on a cross‑validated ranking metric; the threshold that minimized the ranking metrics was selected.  
The application of a sample-specific cluster $p_{cc}$ removed tiny disconnected islands, noise that would otherwise inflate the false positive count.

\noindent\textbf{Threshold identification for label redefinition ($lblredef$):} A second adaptive search fine‑tuned the consistency between labels. 
Jiang \textit{et~al.}~\cite{jiang2024enhancing} suggested a similar approach after the removal of the noisy component. For example, if the enhancing tumor fraction fell below its cluster‑specific cut-off point, ET voxels were relabeled as non‑enhancing core or surrounding edema; an analogous rule was applied to ED when its share of WT was too small. We propose correcting systematic label confusions in a data‑driven way. After removing small components, we built a confusion matrix over all cross‑validated predictions to identify pairs of frequently swapped labels.   For every such pair $(lbl_x,lbl_y)$, we searched, within each cluster, for the cut-off point on the ratio $lbl_x/WT$ that minimizes the ranking metric. If a new case fell below this cutoff point, all $lbl_x$ voxels are converted to $lbl_y$.   This ratio‑based \textit{lblredef} step enforced anatomically plausible label volumes and improved the performance on the BraTS metrics at the lesion level.  

\section{Implementation Details}

For each task, we used the best segmentation models from the previous BraTS edition (see Section~\ref{sec:SM}). These models were trained on BraTS 2024 datasets, which is identical to the current edition BraTS 2025 data. GLI-pre and GLI-post were treated as separate tasks for post-processing (GLI-post included RC). The labels being the same, the SSA task used the same pipeline for hyperparameter search as GLI-pre. The post-processing used the binary labels produced by the segmentation models as input.






\section{Results}

\begin{figure}[htbp]
    \centering
    \includegraphics[width=\linewidth]{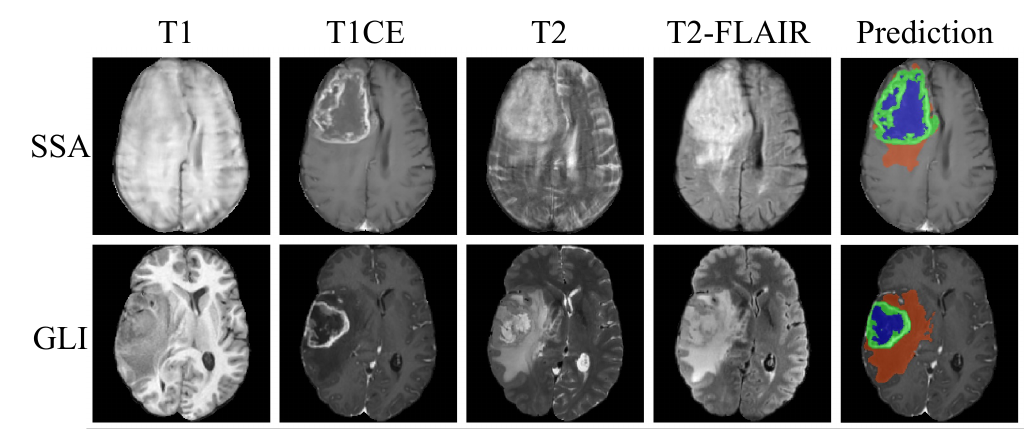}
    \caption{Qualitative results showing median lesion-wise Dice of the whole tumor. SSA: 0.959, orange=ED, blue=NCR, green=ET; GLI: 0.947, orange=SNFH, blue=NETC, green=ET.}
    \label{fig:results}
\end{figure}

Table~\ref{tab:val-results-peds} provides an overview of the performance evaluation of our post-processed models for the validation set of the GLI and SSA tasks. Additionally, results on the testing set are included in Table~\ref{tab:test-results-peds}.  The reported numbers are obtained from the automatic pipeline setup in the BraTS 2025 digital platform with no access to the ground truth of the validation set and no access to any testing data including images and labels. Submission-related CSVs were downloaded from the platform and used to obtain the ranking metric.
Figure~\ref{fig:results} illustrated qualitative results on validation cases for SSA and GLI.

\begin{table}[h]
\caption{\textbf{Quantitative results on the validation subset} of GLI and SSA. Lesion-wise (LW) Dice coefficients and Normalized Surface Distance (NSD) with thresholds of 1.0 mm were computed for enhancing tumor (ET), tumor core (TC), whole tumor (WT), non-enhancing tumor core (NETC), surrounding non-enhancing FLAIR hyper-intensity (SNFH), and resection cavity (RC), respectively. $(\uparrow)$ represents a metric where a higher value is better and $(\downarrow)$ represents a metric where a lower value is better. The best performing ranking metric for each task is highlighted in \textbf{bold}. SM: segmentation model.}
\centering
\resizebox{1.0\textwidth}{!}{%
\begin{tabular}{clcccccccccccccccc}
\hline
\multirow{2}{*}{\textbf{Task}} & \multicolumn{1}{c}{\multirow{2}{*}{\textbf{Model}}} & \multicolumn{1}{c}{\multirow{2}{*}{\textbf{\begin{tabular}[c]{@{}c@{}}GPU\\ Time(hrs)\end{tabular}}}}&\multicolumn{1}{c}{\multirow{2}{*}{\textbf{\begin{tabular}[c]{@{}c@{}}Ranking\\ Metric$(\downarrow)$\end{tabular}}}} & \multicolumn{1}{c}{\textbf{}} & \multicolumn{6}{c}{\textbf{LW Dice}$(\uparrow)$} & \textbf{} & \multicolumn{6}{c}{\textbf{LW NSD}$(\uparrow)$} \\ \cline{6-18} 
 & \multicolumn{1}{c}{} & \multicolumn{1}{c}{} & \multicolumn{1}{c}{\textbf{}} & &\textbf{ET} & \textbf{TC} & \textbf{WT} & \textbf{NETC} & \textbf{SNFH} & \textbf{RC} & \textbf{} & \textbf{ET} & \textbf{TC} & \textbf{WT} & \textbf{NETC} & \textbf{SNFH} & \textbf{RC} \\ \hline
\multirow{3}{*}{\textbf{\begin{tabular}[c]{@{}c@{}}GLI\\ (n = 407)\end{tabular}}} & SM & 401\tablefootnote{estimated based on the segmentation models used; does not include synthetic data generation training } & 1.137 & & 0.794 & 0.798 & 0.881 & 0.756 & 0.823 & 0.858 &  & 0.838 & 0.782 & 0.832 & 0.770 & 0.819 & 0.860 \\
 & SM + $p_{cc}$ & 0 & 1.129 &  & 0.794 & 0.798 & 0.881 & 0.756 & 0.824 & 0.859 &  & 0.838 & 0.782 & 0.832 & 0.770 & 0.819 & 0.862 \\
 & SM + $p_{cc}$ + \textit{lblredef} & 0 & \textbf{1.127}& & 0.794 & 0.798 & 0.881 & 0.757 & 0.824 & 0.859 &  & 0.838 & 0.782 & 0.832 & 0.771 & 0.819 & 0.862 \\ \hline
\multirow{3}{*}{\textbf{\begin{tabular}[c]{@{}c@{}}SSA\\ (n = 35)\end{tabular}}} & SM & 168 &1.729 &  & 0.870 & 0.865 & 0.926 & \multicolumn{3}{c}{\multirow{3}{*}{}} &  & 0.871 & 0.812 & 0.849 & \multicolumn{3}{l}{\multirow{3}{*}{}} \\
 & SM + $p_{cc}$ & 0 &1.629 &  & 0.870 & 0.865 & 0.927 & \multicolumn{3}{c}{} &  & 0.872 & 0.812 & 0.849 & \multicolumn{3}{l}{} \\
 & SM + $p_{cc}$ + \textit{lblredef} & 0 &\textbf{1.471} &  & 0.870 & 0.865 & 0.927 & \multicolumn{3}{c}{} &  & 0.872 & 0.812 & 0.849 & \multicolumn{3}{l}{} \\ \hline
\end{tabular}
}
\label{tab:val-results-peds}
\end{table}

\begin{table}[h]
\caption{\textbf{Quantitative results on the test subset} of GLI and SSA. These results were calculated on a hidden test set and provided by the organizers post-challenge. Lesion-wise (LW) Dice coefficients and Normalized Surface Distance (NSD) with thresholds of 1.0 mm were computed for enhancing tumor (ET), tumor core (TC), whole tumor (WT), non-enhancing tumor core (NETC), surrounding non-enhancing FLAIR hyper-intensity (SNFH), and resection cavity (RC), respectively. $(\uparrow)$ represents a metric where a higher value is better and $(\downarrow)$ represents a metric where a lower value is better. SM: segmentation model.}
\centering
\resizebox{1.0\textwidth}{!}{%
\begin{tabular}{cllcccccccccc}
\hline
\multirow{2}{*}{\textbf{Task}} & &\multicolumn{1}{c}{\multirow{2}{*}{\textbf{Model}}} & \multicolumn{1}{c}{\textbf{}} & \multicolumn{4}{c}{\textbf{LW Dice}$(\uparrow)$} & \textbf{} & \multicolumn{4}{c}{\textbf{LW NSD}$(\uparrow)$} \\ \cline{5-13} 
 & \multicolumn{1}{c}{} & \multicolumn{1}{c}{} &  &\textbf{ET} & \textbf{TC} & \textbf{WT}  & \textbf{RC} & \textbf{} & \textbf{ET} & \textbf{TC} & \textbf{WT} & \textbf{RC} \\ \hline
\textbf{GLI} & &SM + $p_{cc}$ + \textit{lblredef} &  & 0.813 & 0.813 & 0.882 & 0.894 &  & 0.855 & 0.821 & 0.851 &  0.892 \\ \hline
\textbf{SSA} & &SM + $p_{cc}$ + \textit{lblredef} &   & 0.900 & 0.912 & 0.934 & \multicolumn{1}{c}{} &  & 0.904 & 0.865 & 0.872 & \\ \hline
\end{tabular}
}
\label{tab:test-results-peds}
\end{table}

For the GLI task in Table~\ref{tab:val-results-peds}, the lesion‑wise Dice scores increase from 0.756 to 0.757 for NETC, 0.823 to 0.824 for SNFH and 0.858 to 0.859 for RC due to post-processing. Lesion‑wise NSD also improves, from 0.770 to 0.771 for NETC and 0.860 to 0.862 for RC. Together, these refinements reduce the overall ranking metric from 1.137 to 1.127, showing that the post‑processed results perform better than the original segmentation model.

Also, in Table~\ref{tab:val-results-peds} for the SSA task, post-processing improves the lesion-wise Dice of the WT from 0.926 to 0.927 and the lesion-wise NSD for the ET from 0.871 to 0.872. The ranking metric improved from 1.729 to 1.471.

\begin{figure}[htbp]
    \centering
    \begin{subfigure}{0.49\textwidth}
        \includegraphics[width=\linewidth]{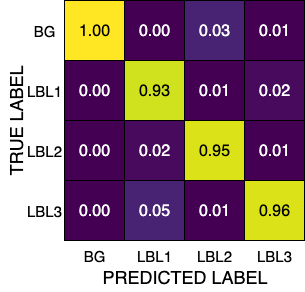}
        \caption{GLI-pre}
        \label{fig:first}
    \end{subfigure}
    \hfill
    \begin{subfigure}{0.49\textwidth}
        \includegraphics[width=\linewidth]{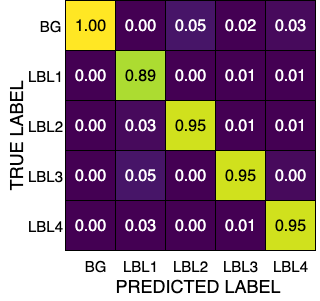}
        \caption{GLI-post}
        \label{fig:second}
    \end{subfigure}
    
    \caption{\textbf{Confusion matrices} illustrating systematic label confusions across cross-validated predictions of GLI-pre and GLI-post after $pp_{cc}$. These confusion matrices are used to identify the labels that are redefined as part of $lblredef$. For example in GLI-pre $lbl_1$ is redefined to $lbl_3$ based on the $lbl_1$/WT ratio.}
    \label{fig:two_subfigs}
\end{figure}

In Figure \ref{fig:first}, we see the confusion matrix for the GLI-pre cross-validated set after the $pp_{cc}$ step. We see that the biggest error is $lbl_3$ falsely predicted as $lbl_1$. Similarly, in Figure \ref{fig:second} for GLI-post the biggest errors are $lbl_3$ falsely predicted as $lbl_1$. These error regions are the focus of $lblredef$, where $lbl_1$/WT are identified as $lbl_1  \rightarrow lbl_3$ for each of the clusters.
\section{Discussion}

For the GLI validation set, adaptive post‑processing nudged metrics only slightly (NETC Dice $+0.001$, SNFH Dice $+0.001$, RC NSD $+0.002$), improving the BraTS ranking score by just $0.9$ \% ($1.137 \rightarrow 1.127$). The ensemble is therefore already near the task ceiling; further gains will likely need (i) stronger anatomical priors for smoother boundaries or (ii) fine‑tuning on the few failure cases.

In the resource‑limited SSA cohort, the Dice scores were unchanged ($\le 0.001$), but the ranking score improved by $14.9$ \% ($1.729  \rightarrow  1.471$). Because this metric penalizes severely poor segmented subjects, radiomics-guided threshold tuning helped improve the segmentation scores for outliers. This could be largely attributed to the poor quality of SSA data acquisition, leading to more false positives in the segmentations from the  pre-trained model. This leads to better result improvement after applying the post-processing strategies. 

Training the full ensemble models costs $401$ GPU hours (GLI) and $168$ GPU hours (SSA). In contrast, the entire post‑processing pipeline—PyRadiomics feature extraction, $k$ means clustering, and a grid search over $p_{cc}$ and $lblredef$ thresholds—ran on CPUs and did not use any GPU time.

It is important to note that the optimization for the tasks was tailored to improve the BraTS ranking score, which does not necessarily translate into clinically better segmentations. For medical relevance, optimizing directly for Dice or NSD, or a combination of both, may be more meaningful than maximizing challenge ranking, as these metrics better capture overlap and boundary accuracy in a clinical setting.

The rules of connected component and volume ratio appear saturated in GLI; integrating shape descriptors, vascular atlases, or uncertainty maps could further smooth out the boundaries. Because radiomics are derived from predicted masks, an iterative loop that alternates segmentation and feature extraction is another avenue for improvement. Further improvement in radiomics extraction can be improved by calculating the radiomics on the WT of a lesion instead of the WT of the entire case. 

Finally, to facilitate reproducibility and extend the utility of our adaptive post-processing approach, we have made the complete pipeline publicly available as easy-to-use Docker containers and a webapp. This enables researchers and clinicians to easily deploy, test, and build on our methods without the need for complex setups. The Docker images are hosted at: \url{https://hub.docker.com/r/aparida12/brats2025} and the webapp is accessible at:  \url{https://segmenter.hope4kids.io/}.

\section{Conclusion}
We demonstrated that an adaptive, radiomic-based postprocessing pipeline improved the accuracy of brain segmentation models on multiple BraTS 2025 glioma segmentation tasks with zero additional GPU hours. These findings promote shifting the focus from building increasingly larger models or further resource intensive training schemes, towards data‑driven efficient post-processing strategies.  By fine-tuning large-scale segmentation models using adaptive post-processing, we can make automated brain tumor segmentation both accurate and equitable.

\begin{credits}
\subsubsection{\ackname} This work was supported by the National Cancer Institute (UG3 CA236536), the Spanish  Ministerio de Ciencia e Innovación, the Agencia Estatal de Investigación, NextGenerationEU grants PDC2022-133865-I00 and PID2022-141493OB-I00, and the EUCAIM project co-funded by the European Union (Grant Agreement \#101100633). The authors acknowledge the Universidad Politécnica de Madrid for providing computing resources on the Magerit Supercomputer. This work was also supported by the Comunidad de Madrid, Spain through the MAGERIT-CM project (TEC-2024/COM-44).

\end{credits}

%
%
%
\bibliographystyle{splncs04}
\bibliography{refrences}

\end{document}